\def\year{2022}\relax
\documentclass[letterpaper]{article} 
\usepackage{aaai22}  
\usepackage{times}  
\usepackage{helvet}  
\usepackage{courier}  
\usepackage[hyphens]{url}  
\usepackage{graphicx} 
\usepackage{color, colortbl}
\definecolor{Gray}{gray}{0.9}
\usepackage{natbib}  
\usepackage{caption} 
\DeclareCaptionStyle{ruled}{labelfont=normalfont,labelsep=colon,strut=off} 
\usepackage{amsmath}
\usepackage[linesnumbered,ruled]{algorithm2e}
\usepackage{xcolor}
\usepackage{pifont}
\usepackage{amssymb}
\newcommand{\proposed}{\textsf{AFGRL}}
\newcommand{\cmark}{\textcolor{blue}{\ding{51}}}%
\newcommand{\xmark}{\textcolor{red}{\ding{55}}}%
\usepackage{multirow}
\usepackage{setspace}
\usepackage[hidelinks]{hyperref}

\title{Augmentation-Free Self-Supervised Learning on Graphs}
\author {
    Namkyeong Lee\textsuperscript{\rm 1},
    Junseok Lee\textsuperscript{\rm 1},
    Chanyoung Park\textsuperscript{\rm 1,2}\footnote{Corresponding author.}
}
\affiliations {
    \textsuperscript{\rm 1} Dept. of Industrial and Systems Engineering, KAIST, Daejeon, Republic of Korea\\
    \textsuperscript{\rm 2} Graduate School of Artificial Intelligence, KAIST, Daejeon, Republic of Korea\\
    namkyeong96@kaist.ac.kr, junseoklee@kaist.ac.kr, cy.park@kaist.ac.kr
}

\begin{document}

\maketitle

\begin{abstract}
Inspired by the recent success of self-supervised methods applied on images, self-supervised learning on graph structured data has seen rapid growth especially centered on augmentation-based contrastive methods. However, we argue that without carefully designed augmentation techniques, augmentations on graphs may behave arbitrarily in that the underlying semantics of graphs can drastically change. As a consequence, the performance of existing augmentation-based methods is highly dependent on the choice of augmentation scheme, i.e., hyperparameters associated with augmentations.
In this paper, we propose a novel augmentation-free self-supervised learning framework for graphs, named~\proposed. Specifically, we generate an alternative view of a graph by discovering nodes that share the local structural information and the global semantics with the graph. 
Extensive experiments towards various node-level tasks, i.e., node classification, clustering, and similarity search on various real-world datasets demonstrate the superiority of~\proposed. 
The source code for~\proposed~is available at~\url{https://github.com/Namkyeong/AFGRL}.
\end{abstract}

\section{Introduction}

Recently, self-supervised learning paradigm~\cite{liu2021self}, which learns representation from supervision signals derived from the data itself without relying on human-provided labels, achieved great success in many domains including computer vision \cite{rotation, jigsaw}, signal processing \cite{banville2021uncovering,banville2019self}, and natural language processing \cite{BERT, GPT3}.
Specifically, contrastive methods, which are at the core of self-supervised learning paradigm, aim to build effective representation by pulling semantically similar (positive) pairs together and pushing dissimilar (negative) pairs apart~\cite{DeepInfomax, CPC}, where two augmented versions of an image are considered as positives, and the remaining images are considered as negatives.
Inspired by the success of the contrastive methods in computer vision applied on images, these methods have been recently adopted to graphs~\cite{xie2021self}.


\begin{figure}[t]
\centering
\includegraphics[width=0.99\columnwidth]{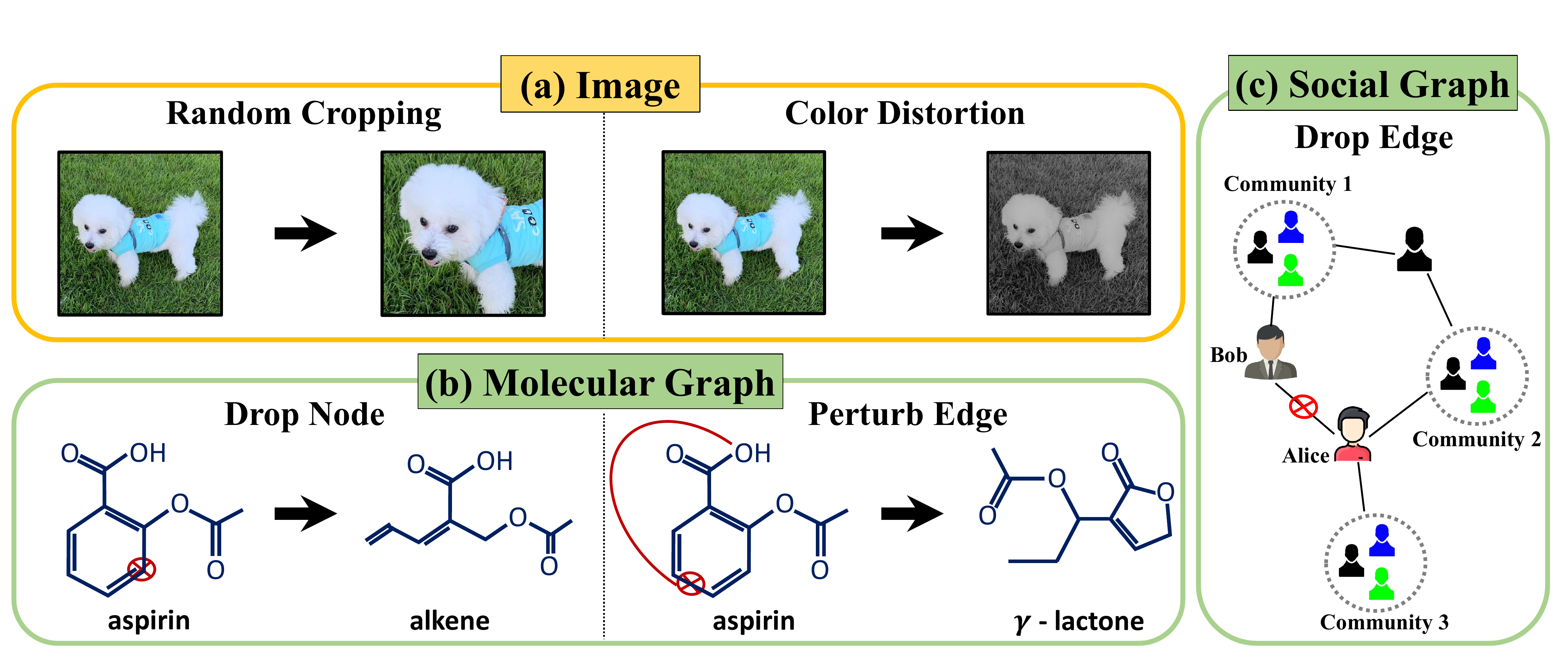} 
\vspace{-1ex}
\caption{Augmentations on images ((a)) keep the underlying semantics, whereas augmentations on graphs ((b),(c)) may unexpectedly change the semantics.}
\label{fig1}
\vspace{-2ex}
\end{figure}

Although self-supervised contrastive methods have been shown to be effective on various graph-related tasks, they pay little attention to the inherent distinction between images and graphs: 
\textit{while augmentation is well defined on images, it may behave arbitrarily on graphs. }
For example, in the case of images, even after randomly cropping and rotating them, or distorting their color, their underlying semantic is hardly changed (Figure~\ref{fig1} (a)), and even if the semantic changes, humans can readily recognize the change visually, and choose an alternative augmentation approach that preserves the semantic. 
On the other hand, when we perturb (drop or add) edges/nodes, and their features of a graph, we cannot ascertain whether the augmented graph would be positively related to the original graph, and what is worse, it is non-trivial to verify the validity of the augmented graph since graphs are hard to visualize. 
For example, in molecular graphs, dropping a carbon atom from the phenyl ring of aspirin breaks the aromatic system and results in a alkene chain (Figure~\ref{fig1}(b)). Moreover, perturbing the connection of aspirin might introduce a molecule of totally different property, namely, five-membered lactone~\cite{MoCL}. Likewise, in social graphs, randomly dropping edges might also lead to semantic changes, especially when these edges are related to hub nodes. For example, as shown in Figure~\ref{fig1}(c), if the edge between Bob and Alice is dropped, it would take much longer distance for Bob to reach Community 3 (i.e., from 2-hops to 5-hops), which also alters the relationship between Community 1 and Community 3.
We argue that this is mainly because graphs contain not only the semantic but also the \textit{structural information}.

Due to the aforementioned arbitrary behavior of augmentation on graphs, the quality of the learned graph representations of previous augmentation-based contrastive methods~\cite{MVGRL,GRACE,GCA,BGRL,MoCL,DGI} is highly \textbf{dependent on the choice of the augmentation scheme}.
More precisely, in order to augment graphs, these methods perform various augmentation techniques such as node/edge perturbation and node feature masking, and the amount by which the graphs are augmented is controlled by a set of hyperparameters. However, these hyperparameters should be carefully tuned according to which datasets, and which downstream tasks are used for the model evaluation, otherwise the model performance would vary greatly~\cite{you2021graph}. Moreover, it is also shown that the performance on downstream tasks highly resort to which combinations of the augmentation techniques~\cite{GraphCL, MoCL} are used.

Furthermore, even after discovering the best hyperparameters for augmentations, another limitation arises due to the inherent philosophy of contrastive learning. More precisely, inheriting the principle of instance discrimination~\cite{IND}, contrastive methods treat two samples as a positive pair as long as they are two augmented versions of the same instance, and all other pairs are treated as negatives. Although this approach is effective for learning representations of images~\cite{SimCLR,MoCo}, simply adopting it to graphs by \textbf{treating all other nodes apart from the node itself as negatives overlooks the structural information }of graphs, and thus cannot benefit from the relational inductive bias of graph-structured data~\cite{battaglia2018relational}. 
Lastly, due to the nature of contrastive methods, \textbf{a large amount of negative samples} is required for improving the performance on the downstream tasks, requiring high computational and memory costs, which is impractical in reality~\cite{BGRL}.

\smallskip
\noindent\textbf{Contribution}. 
We propose a self-supervised learning framework for graphs, called \textsf{A}ugmentation-\textsf{F}ree \textsf{G}raph \textsf{R}epresentation \textsf{L}earning (\proposed), which \textit{requires neither augmentation techniques nor negative samples} for learning representations of graphs.
Precisely, instead of creating two arbitrarily augmented views of a graph and expecting them to preserve the semantics of the original graph, we use the original graph per se as one view, and generate another view by discovering, for each node in the original graph, nodes that can serve as \textit{positive samples} via $k$-nearest-neighbor ($k$-NN) search in the representation space. Then, given the two semantically related views, we aim to predict, for each node in the first view, the latent representations of its positive nodes in the second view. However, naively selecting positive samples based on $k$-NN search to generate an alternative view can still alter the semantics of the original graph. 

Hence, we introduce a mechanism to filter out false positives from the samples discovered by $k$-NN search. In a nutshell, we consider a sample to be positive only if either 1) it is a neighboring node of the target node in the adjacency matrix (local perspective), capturing the relational inductive bias inherent in the graph-structured data, or 2) it belongs to the same cluster as the target node (global perspective). Moreover, by adopting BYOL~\cite{BYOL} as the backbone of our model, negative samples are not required for the model training, thereby avoiding the \textit{``sampling bias''}~\cite{lin2021prototypical}, i.e. the negative samples may have the same semantics with the query node, which would result in less effective representation~\cite{saunshi2019theoretical}.


Our extensive experiments demonstrate that~\proposed~outperforms a wide range of state-of-the-art methods in terms of node classification, clustering and similarity search. We also demonstrate that compared with existing methods that heavily depend on the choice of hyperparameters,
~\proposed~is stable over hyperparameters. To the best of our knowledge,~\proposed~is the first work that learns representations of graphs without relying on manual augmentation techniques and negative samples.

\section{Related Work}

\noindent\textbf{Contrastive Methods on Graphs. }
Recently, motivated by the great success of self-supervised methods on images, contrastive methods have been increasingly adopted to graphs. 
DGI~\cite{DGI}, a pioneering work highly inspired by Deep InfoMax~\cite{DeepInfomax}, aims to learn node representations by maximizing the mutual information between the local patch of a graph. i.e., node, and the global summary of the graph, thereby capturing the global information of the graph that is overlooked by vanilla graph convolutional networks (GCNs)~\cite{GCN,GAT}.
DGI is further improved by taking into account the mutual information regarding the edges~\cite{GMI} and node attributes~\cite{jing2021hdmi}. 
Inspired by SimCLR~\cite{SimCLR}, GRACE \cite{GRACE} first creates two \textit{augmented views} of a graph by randomly perturbing nodes/edges and their features. Then, it learns node representations by pulling together the representation of the same node in the two augmented graphs, while pushing apart representations of every other node.
This principle~\cite{IND} has also been adopted for learning graph-level representations of graphs that can be used for graph classification,~\cite{infograph,GraphCL,MVGRL}.
Despite the success of contrastive methods on graphs, they are criticized for the problem raised by the ``sampling bias''~\cite{bielak2021graph}.
Moreover, these methods require a large amount of negative samples for the model training, which incurs high computational and memory costs~\cite{BYOL}.

To address the sampling bias issue, BGRL \cite{BGRL} learns node representations without using negative samples. It learns node representations by encoding two augmented versions of a graph using two separate encoders: one is trained through minimizing the cosine loss between the representations generated by the two encoders, while the other encoder is updated by an exponential moving average of the first encoder. 
Although the sampling bias has been addressed in this way, BGRL still requires augmentations on the original graph, which may lead to semantic drift~\cite{MoCL} as illustrated in Figure~\ref{fig1}. On the other hand, our proposed method learns node representations without any use of negative samples or augmentations of graphs.

\smallskip
\noindent\textbf{Augmentations on Graphs. }
Most recently, various augmentation techniques for graphs have been introduced. e.g., node dropping \cite{GraphCL}, edge modification \cite{jin2021multi, qiu2020gcc, zhao2020data}, subgraph extraction \cite{subg-con, sun2021sugar}, attribute masking \cite{GRACE, GCA} and others \cite{MVGRL, kefato2021self, suresh2021adversarial}. GRACE \cite{GRACE} randomly drops edges and masks node features to generate two augmented views of a graph. 
GCA \cite{GCA} further improves GRACE by introducing advanced adaptive augmentation techniques that take into account both structural and attribute information. However, due to the complex nature of graphs, the performance on downstream tasks is highly dependent on the selection of the augmentation scheme, as will be shown later in our experiments (Table \ref{tab:augmentation_sensitivity}). Moreover, previous work~\cite{GraphCL, MoCL} have shown that there is no universally outperforming data augmentation scheme for graphs.
Lastly,~\citet{MoCL} demonstrates that infusing domain knowledge is helpful in finding proper augmentations, which preserves biological assumption in molecular graph. However, domain knowledge is not always available in reality. In this work, we propose a general framework for generating an alternative view of the original graph without relying on existing augmentation techniques that may either 1) change the semantics of the original graph or 2) require domain knowledge.

\section{Problem Statement}
\subsubsection{Notations.} 
Let $\mathcal{G} = (\mathcal{V}, \mathcal{E})$ denote a graph, where $\mathcal{V} = \left \{v_1, ... , v_N \right \}$ represents the set of nodes, and ~$\mathcal{E} \subseteq \mathcal{V} \times \mathcal{V} $ represents the set of edges. $\mathcal{G}$ is associated with a feature matrix $\mathbf{X} \in \mathbb{R}^{N \times F}$, and an adjacency matrix $\mathbf{A} \in \mathbb{R}^{N \times N}$ where $\mathbf{A}_{ij} = 1$ iff $(v_i, v_j) \in \mathcal{E}$ and $\mathbf{A}_{ij} = 0$ otherwise. 
\subsubsection{Task: Unsupervised Graph Representation Learning.} Given a graph $\mathcal{G}$ along with $\mathbf{X}$ and $\mathbf{A}$, we aim to learn a encoder $f(\cdot)$ that produces node embeddings $\mathbf{H} = f(\mathbf{X}, \mathbf{A})\in\mathbb{R}^{N\times D}$, where $D<<F$. In particular, our goal is to learn node embeddings that generalize well to various downstream tasks
without using any class information.

\section{Preliminary: Bootstrap Your Own Latent}
\label{sec:BYOL}
Before explaining details of our proposed method, we begin by introducing BYOL~\cite{BYOL}, which is the backbone of our proposed framework.
The core idea of BYOL is to learn representations of images without using negative samples~\cite{BYOL}.
Given two augmented views of an image, BYOL trains two separate encoders, i.e., online encoder $f_{\theta}$ and target encoder $f_{\xi}$, and learns representations of images by maximizing the similarity of the two representations produced by each encoder. More formally,
BYOL generates two views $\mathbf{{x}}_1 \sim {t}(\mathbf{x})$, and $\mathbf{{x}}_2 \sim {t'}(\mathbf{x})$ of an image $\mathbf{x}$ given a set of transformations $t\sim\mathcal{T}$ and $t'\sim\mathcal{T}$, and these two generated views of an image are fed into the online and target encoders. Precisely, the online encoder  $f_{\theta}$ produces online representation $\mathbf{{h}}_1 = f_{\theta}(\mathbf{{x}}_1)$, while the target encoder  $f_{\xi}$ produces target representation $\mathbf{{h}}_2 = f_{\xi}(\mathbf{{x}}_2)$. Then, both online and target representations are projected to smaller representations $\mathbf{{z}}_1=g_{\theta}(\mathbf{{h}}_1)$ and $\mathbf{{z}}_2=g_{\xi}(\mathbf{{h}}_2)$ using projectors $g_{\theta}$ and $g_{\xi}$, respectively. 
Finally, an additional predictor $q_{\theta}$ is applied on top of the projected online representation, i.e., $\mathbf{{z}_1}$, to make the architecture asymmetric. The objective function is defined as $\mathcal{L_{\theta, \xi}} = {\left \|{\bar{q}_{\theta}}(\mathbf{{z}_1}) - {\bar{\mathbf{z}}_2} \right \|}^2$,
where $\bar{q}_{\theta}(\mathbf{{z}_1})$ and $\bar{\mathbf{z}}_2$ denote $l2$-normalized form of ${q}_{\theta}(\mathbf{{z}_1})$ and ${\mathbf{z}}_2$, respectively.  A symmetric loss $\widetilde{\mathcal{L}}_{\theta, \xi}$ is obtained by feeding $\mathbf{{x}_2}$ into the online encoder and $\mathbf{{x}_1}$ into the target encoder, and the final objective is to minimize $\mathcal{L}_{\theta, \xi}^{\mathrm{BYOL}}=\mathcal{L}_{\theta, \xi}+\widetilde{\mathcal{L}}_{\theta, \xi}$. At each training iteration, a stochastic optimization step is performed to minimize $\mathcal{L}_{\theta, \xi}^{\mathrm{BYOL}}$ with respect to $\theta$ only, while $\xi$ is updated using the exponential moving average (EMA) of $\theta$, which is empirically shown to prevent the collapsing problem \cite{SimSiam}. More formally, the parameters of BYOL are updated as $\theta \leftarrow \operatorname{optimizer}\left(\theta, \nabla_{\theta} \mathcal{L}_{\theta, \xi}^{\mathrm{BYOL}}, \eta\right),~~ \xi\leftarrow\tau\xi+(1-\tau)\theta$,
where $\eta$ is learning rate for online network, and $\tau \in [0, 1]$ is the decay rate that controls how close $\xi$ remains to $\theta$. 

\begin{figure*}
	\centering
	\includegraphics[width=0.85\linewidth]{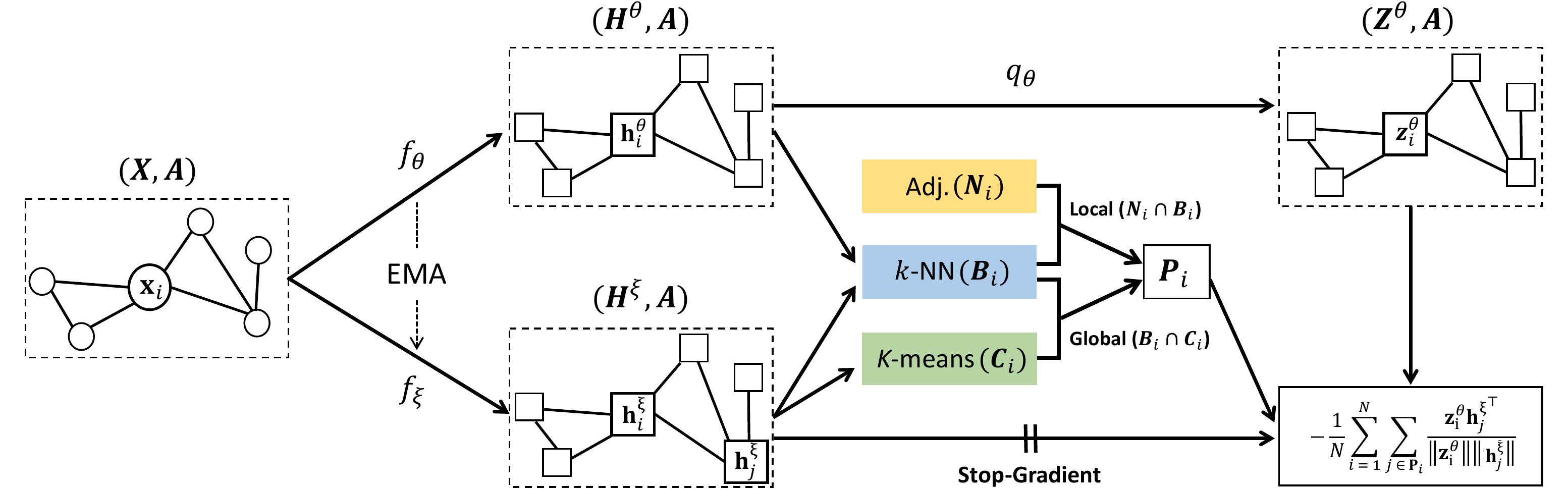} 
	\caption{The overall architecture of~\proposed. Given a graph, $f_\theta$ and $f_\xi$ generate node embeddings $\mathbf{H}^\theta$ and $\mathbf{H}^\xi$ both of which are used to obtain $k$-NNs for node $v_i$, i.e., $\mathbf{B}_i$. Combining it with $\mathbf{N}_i$, we obtain local positives, i.e., $\mathbf{B}_i\cap\mathbf{N}_i$. 
	To obtain global positives for node $v_i$, $K$-means clustering is performed on $\mathbf{H}^\xi$, and the result $\mathbf{C}_i$ is combined with $\mathbf{B}_i$, i.e., $\mathbf{B}_i\cap\mathbf{C}_i$. Finally, we combine local and global positives to obtain real positives, i.e., $\mathbf{P}_i$. 
	A predictor ${q}_\theta$ projects $\mathbf{H}^\theta$ to $\mathbf{Z}^\theta$, which is used to compute the final loss along with $\mathbf{H}^\xi$. Note that $f_\theta$ is updated via gradient descent of the loss, whereas $f_\xi$ is updated via EMA of $f_\theta$.} 
	\label{exp:ablation_clustering}
	\vspace{-3ex}
\end{figure*}

\section{Proposed Method}

We first introduce how BYOL has been previously employed on graphs~\cite{BGRL}, and discuss about the several limitations of augmentation-based methods for graphs.
Finally, we present our proposed method, called~\proposed.


\smallskip
\noindent\textbf{Generating Alternative Views via Augmentation.}
BGRL~\cite{BGRL} is a recently proposed fully non-contrastive method for learning node representations that does not leverage negative samples benefiting from the framework of BYOL~\cite{BYOL}. Precisely, BGRL generates two different views of a graph via manual augmentations, i.e., {node feature masking} and {edge masking} as done by previous methods \cite{GRACE, GCA}, and the amount by which the graphs are augmented is controlled by a set of hyperparameters. Then, two encoders, i.e., online and target encoders, generate embeddings given the augmented views of a graph as inputs, and the two generated embeddings are learned to be close to each other. To prevent the representations from collapsing to trivial solutions, BGRL introduces a symmetry-breaking technique (refer to Section~\ref{sec:BYOL} for more detailed explanation). It is also worth noting that BGRL intentionally considers simple augmentation techniques to validate the benefit of fully non-contrastive scheme applied on graphs.

\smallskip
\noindent\textbf{Limitation of Augmentation-based Methods on Graphs.}
Although BGRL has been shown to be effective in a fully non-contrastive manner, i.e., without using negative samples, we observe that \textit{the quality of the learned node representations relies on the choice of the augmentation scheme}.  In other words, performance on various downstream tasks evaluated based on the representations learned by BGRL varies greatly according to the choice of hyperparameters associated with augmentations, and the best hyperparameters are different for different datasets. This phenomenon becomes even clearer when stronger augmentations, such as diffusion~\cite{MVGRL}, adaptive techniques~\cite{GCA} and the infusion of domain knowledge~\cite{MoCL} are applied. 
Table~\ref{tab:augmentation_sensitivity} shows how the performance of augmentation-based methods varies according to the hyperparameters associated with augmentations. More precisely, we report the relative performance of the best performing case compared to the worst performing case, i.e., -4.00\% indicates that the worst case performs 4\% worse than the best case. We observe that the performance in both tasks is sensitive to the hyperparameters, and that it aggravates when a stronger augmentation technique is employed, i.e., GCA, in which case the role of augmentation becomes even more important.
Thus, we need a more stable and general framework for generating an alternative view of the original graph without relying on augmentation techniques introduced in existing works.

\begin{table}[t]{
					\centering
			{\small
				\begin{tabular}{p{0.7cm}|p{0.7cm}|cccc}
					\noalign{\smallskip}\noalign{\smallskip}
					\multicolumn{2}{c|}{} & Comp. & Photo & CS & Physics \\
					\hline
					\multirow{2}{*}{\begin{tabular}[x]{@{}c@{}}Node\\Classi.\end{tabular}} & BGRL & -4.00\% & -1.06\% & -0.20\% & -0.69\% \\
					& GCA & -19.18\% & -5.48\% & -0.27\% & OOM \\
					\hline
					\multirow{2}{*}{\begin{tabular}[x]{@{}c@{}}Node\\Clust.\end{tabular}} & BGRL & -11.57\% & -13.30\% & -0.78\% & -6.46\% \\
					& GCA & -26.28\% & -23.27\% & -1.64\% & OOM \\
				\end{tabular}
			}
		\vspace{-1ex}
			\caption{Performance sensitivity according to the hyperparameters for augmentations (i.e., edge drop and node feature masking) on node classification and clustering. Each value indicates the relative performance difference between the best vs. worst performing cases, i.e. $-\frac{(\text{best}-\text{worst})}{\text{best}} \times 100$. The hyperparameters (i.e., probability of edge drop and node feature masking) are chosen within the range from 0.0 to 0.5 to prevent a significant distortion of graphs.}
			\label{tab:augmentation_sensitivity}
		}
\vspace{-2ex}
\end{table}%

\subsection{Augmentation-Free GRL (\proposed)}
We propose a simple yet effective self-supervised learning framework for generating an alternative view of the original graph taking into account the relational inductive bias of graph-structured data, and the global semantics of graphs.
For each node $v_i\in\mathcal{V}$ in graph $\mathcal{G}$, we discover nodes that can serve as positive samples based on the node representations learned by the two encoders. i.e., online encoder $f_\theta(\cdot)$ and target encoder $f_\xi(\cdot)$\footnote{\proposed~adopts the architecture of BGRL~\cite{BGRL}, which is slightly modified from BYOL~\cite{BYOL}. In particular, projection networks, i.e., $g_\theta(\cdot)$ and $g_\xi(\cdot)$ are not used.}.
More precisely, these encoders initially receive the adjacency matrix $\mathbf{A}$ and the feature matrix $\mathbf{X}$ of the original graph as inputs, and compute the online and target representations. i.e., $\mathbf{H}^\theta=f_\theta(\mathbf{X}, \mathbf{A})$ and $\mathbf{H}^\xi=f_\xi(\mathbf{X}, \mathbf{A})$ whose $i$-th rows, $\mathbf{h}_i^{\theta}$ and $\mathbf{h}_i^{\xi}$, are representations for node $v_i\in\mathcal{V}$. Then, for a given query node $v_i\in\mathcal{V}$, we compute the cosine similarity between all other nodes in the graph as follows:
\vspace{-1ex}
\begin{eqnarray}
\small
	sim(v_i, v_j) = \frac{{\mathbf{h}}_i^{\theta}\cdot{\mathbf{h}}_j^{\xi}}{\|\mathbf{h}_i^\theta\|\|\mathbf{h}_j^\xi\|}, \forall v_j\in\mathcal{V}
\end{eqnarray}
where the similarity is computed between the online and the target representations. Given the similarity information, we search for $k$-nearest-neighbors for each node $v_i$, and denote them by a set $\mathbf{B}_i$, which can serve as positive samples for node $v_i$.
Essentially, we expect the nearest neighbors in the representation space to belong to the same semantic class as the query node, i.e., $v_i$ in this case.
Although $\mathbf{B}_i$ can serve as a reasonable set of positive candidates for node $v_i$, 1) \textit{it is inherently noisy} as we do not leverage any label information, i.e., $\mathbf{B}_i$ contains samples that are not semantically related to the query node $v_i$. Moreover, only resorting to the nearest neighbors in the representation space may not only overlook 2)  \textit{the structural information inherent in the graph, i.e., relational inductive bias}, but also 3) \textit{the global semantics of the graph}. 
To address these limitations, we introduce a mechanism to filter out false positives from the samples discovered by $k$-NN search, while also capturing the local structural information and the global semantics of graphs.

\begin{figure}[t]
	\centering
	\includegraphics[width=0.81\columnwidth]{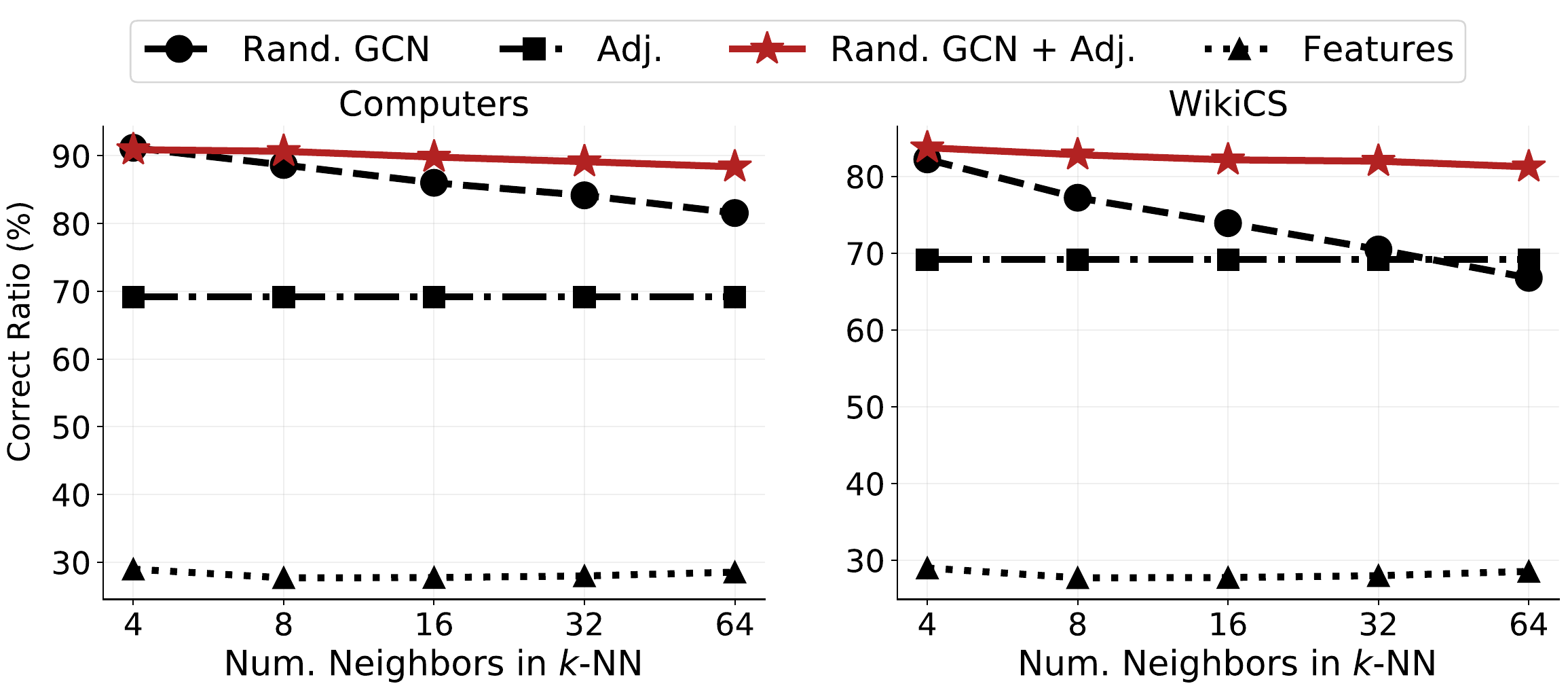}
	\vspace{-1.5ex}
	\caption{Analysis on the ratio of its neighboring nodes being the same label as the query node across different $k$s.}
	\label{fig:ratio}
	\vspace{-4ex}
\end{figure}

\smallskip
\noindent\textbf{Capturing Local Structural Information.}
Recall that we expected the nearest neighbors found by $k$-NN search, i.e., $\mathbf{B}_i$, to share the same class label as the query node $v_i$. To verify whether our expectation holds, we perform analysis on two datasets, i.e., Amazon Computers and WikiCS datasets as shown in Figure~\ref{fig:ratio}.
First, we obtain node embeddings from  a randomly initialized 2-layer GCN ~\cite{GCN}, i.e., $\mathbf{H}_{\textsf{Rand-GCN}}=\textsf{Rand-GCN}(\mathbf{X},\mathbf{A})$, and perform $k$-NN search for each node given the node embeddings $\mathbf{H}_{\textsf{Rand-GCN}}$.
Then, for each node, we compute the ratio of its neighboring nodes being the same label as the query node. In Figure~\ref{fig:ratio}, we observe that although the ratio is high when considering only a small number of neighbors, e.g., $k=4$, the ratio decreases as $k$ gets larger in both datasets. 
This implies that although our expectation holds to some extent, there still exists noise.

Hence, to filter out false positives from the nearest neighbors found by $k$-NN search, i.e., $\mathbf{B}_i$ for each node $v_i$, we leverage the local structural information among nodes given in the form of an adjacency matrix. i.e., relational inductive bias.
More precisely, for a node $v_i$, its adjacent nodes $\mathbf{N}_i$ tend to share the same label as the query node $v_i$, i.e., smoothness assumption~\cite{zhu2003semi}. In Figure~\ref{fig:ratio}, we indeed observe that the ratio of the adjacent nodes being the same label as the query node (``\textit{Adj}'') is about 70\% in both datasets, which demonstrates the validity of the smoothness assumption. 
Therefore, to capture the relational inductive bias reflected in the smoothness assumption, while filtering out false positives from noisy nearest neighbors,
we compute the intersection between the nearest neighbors and adjacent nodes, i.e., $\mathbf{B}_i\cap\mathbf{N}_i$. We denote the set of these intersecting nodes as \textit{\textbf{local positives}} of $v_i$. Indeed, Figure~\ref{fig:ratio} shows that the local positives (``\textit{Rand. GCN + Adj}'') consistently maintain high correct ratio even when $k$ increases.



\smallskip
\noindent\textbf{Capturing Global Semantics.}
To capture the semantics of nodes in a global perspective, we leverage clustering techniques. The intuition is to discover non-adjacent nodes that share the global semantic information with the query node. 
For example, in an academic collaboration network whose nodes denote authors and edges denote collaboration between authors, even though two authors work on the same research topic (i.e., same label), they may not be connected in the graph since they neither collaborated in the past nor share any collaborators. We argue that such semantically similar entities that do not share an edge can be discovered via clustering in a global perspective.
In this regard, we apply $K$-means clustering algorithm on the target representation $\mathbf{H}^{\xi}$ to cluster nodes into a set of $K$ clusters, i.e. $\mathbf{G} = \left \{ G_1, G_2, ... , G_K \right \}$,
and $c(\mathbf{h}_i^\xi)\in\{1,...,K\}$ denotes the cluster assignment of $\mathbf{h}_i^\xi$, i.e., $v_i \in G_{c(\mathbf{h}_i^\xi)}$. Then, we consider the set of nodes that belong to the same cluster as $v_i$, i.e., $\mathbf{C}_i=\{v_j|v_j\in G_{c(\mathbf{h}_i^\xi)}\}$, as its semantically similar nodes in the global perspective. Finally, we obtain the intersection between the nearest neighbors and the semantically similar nodes in the global perspective, i.e., $\mathbf{B}_i \cap \mathbf{C}_i$, and we denote the set of these intersecting nodes as \textit{\textbf{global positives}} of $v_i$.
In other words, nodes that are among the nearest neighbors of $v_i$ and at the same time belong to the same cluster as $v_i$ are considered as globally positive neighbors.
It is important to note that as $K$-means clustering algorithm is sensitive to the cluster centroid initialization, we perform multiple runs to ensure robustness of the clustering results.
Specifically, we perform $K$-means clustering $M$ times and obtain $M$ sets of clusters, i.e., $\{\mathbf{G}^{(j)}\}_{j=1}^M$, where $\mathbf{G}^{(j)} = \left \{ G_1^{(j)}, G_2^{(j)}, ... , G_K^{(j)} \right \}$ is the result of $j$-th run of the clustering. Then, we define $\mathbf{C}_i = \bigcup_{j=1}^{M}G_{c^{(j)}(\mathbf{h}_i^\xi)}^{(j)}$, where $c^{(j)}(\mathbf{h}_i^\xi)\in\{1,...,K\}$ denotes the cluster assignment of $\mathbf{h}_i^{\xi}$ in the $j$-th run of clustering.

\begin{figure}[t]
	\centering
	\includegraphics[width=0.8\columnwidth]{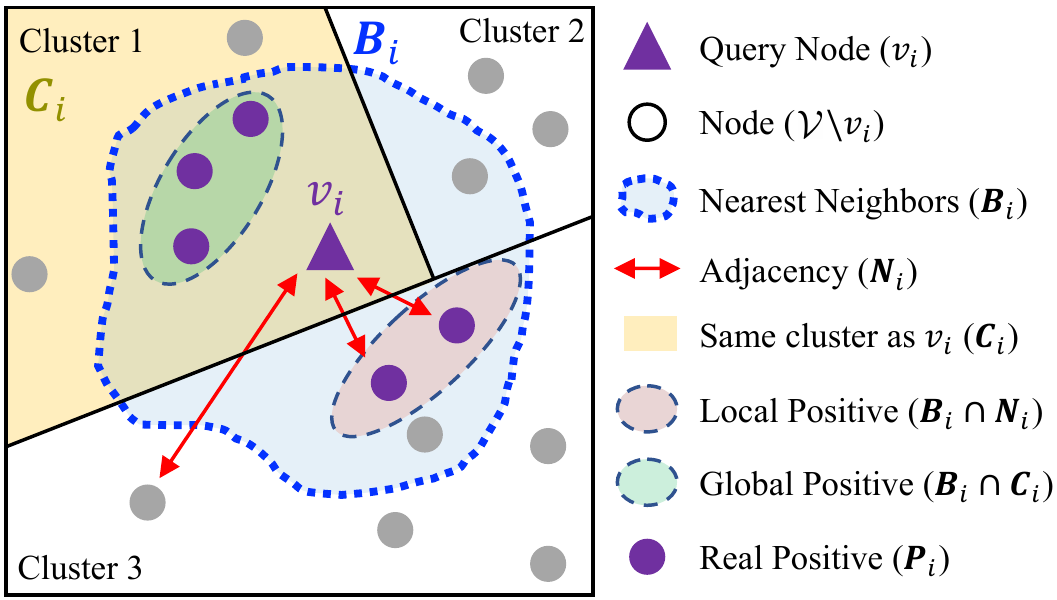} 
	\vspace{-1ex}
	\caption{An overview of obtaining real positives of node $v_i$.}
	\label{fig:clustering}
	\vspace{-3ex}
\end{figure}

\smallskip
\noindent\textbf{Objective Function.}
In order to consider both the local and global information, we define the set of \textit{\textbf{real positives}} for node $v_i$ as follows: 
\begin{equation}
\mathbf{P}_i = (\mathbf{B}_i \cap \mathbf{N}_i) \cup (\mathbf{B}_i \cap \mathbf{C}_i)    
\end{equation}
Our objective function aims to minimize the cosine distance between the query node $v_i$ and its real positives $\mathbf{P}_i$:
\begin{equation}
    \mathcal{L_{\theta, \xi}} = -\frac{1}{N}\sum_{i=1}^{N}\sum_{v_j \in \mathbf{P}_i}{\frac{\mathbf{z}_{i}^{\theta}\mathbf{h}_{j}^{\xi\top} }{\left \| \mathbf{z}_{i}^{\theta} \right \| \left \| \mathbf{h}_{j}^{\xi} \right \|}} ,
    \label{eqn:loss}
\end{equation}
where $\mathbf{z}_i^\theta=q_\theta(\mathbf{h}_i^\theta)\in\mathbb{R}^D$ is the prediction of the online embedding $\mathbf{h}_i^\theta\in\mathbb{R}^D$, and $q_\theta(\cdot)$ is the predictor network. 
Following BYOL,~\proposed's online network is updated based on the gradient of its parameters with respect to the loss function (Equation~\ref{eqn:loss}), while the target network is updated by smoothing the online network. We also symmetrize the loss function. In the end, the online embeddings, i.e., $\mathbf{H}^\theta\in\mathbb{R}^{N\times D}$ are used for downstream tasks. 
Figure~\ref{fig:clustering} illustrates the overview of obtaining real positives of node $v_i$. 

In summary, 1)~\proposed~does not rely on arbitrary augmentation techniques for the model training, thereby achieving stable performance. 2)~\proposed~filters out false positives from the samples discovered by $k$-NN search, while also capturing the local structural information, i.e., relational inductive bias, and the global semantics of graphs. 3)~\proposed~does not require negative samples for the model training, thereby avoiding sampling bias and alleviating computational/memory costs suffered by previous contrastive methods.


\begin{table*}[t]
    \begin{minipage}{.70\linewidth}{
        \centering
        {\small
        				\renewcommand{\arraystretch}{0.75}
        \begin{tabular}{c|ccccc}
        	
            &                WikiCS    & {\footnotesize Computers} & \footnotesize Photo  & Co.CS       & Co.Physics  \\ \hline \hline
                            Sup. GCN        & 77.19 \scriptsize{$\pm$ 0.12} & 86.51 \scriptsize{$\pm$ 0.54} & 92.42 \scriptsize{$\pm$ 0.22} & 93.03 \scriptsize{$\pm$ 0.31} & 95.65 \scriptsize{$\pm$ 0.16} \\ \hline 
                Raw feats.        & 71.98 \scriptsize{$\pm$ 0.00} & 73.81 \scriptsize{$\pm$ 0.00} & 78.53 \scriptsize{$\pm$ 0.00} & 90.37 \scriptsize{$\pm$ 0.00} & 93.58 \scriptsize{$\pm$ 0.00} \\
                node2vec            & 71.79 \scriptsize{$\pm$ 0.05} & 84.39 \scriptsize{$\pm$ 0.08} & 89.67 \scriptsize{$\pm$ 0.12} & 85.08 \scriptsize{$\pm$ 0.03} & 91.19 \scriptsize{$\pm$ 0.04} \\
                DeepWalk            & 74.35 \scriptsize{$\pm$ 0.06} & 85.68 \scriptsize{$\pm$ 0.06} & 89.44 \scriptsize{$\pm$ 0.11} & 84.61 \scriptsize{$\pm$ 0.22} & 91.77 \scriptsize{$\pm$ 0.15} \\
                DW + feats. & 77.21 \scriptsize{$\pm$ 0.03}  & 86.28 \scriptsize{$\pm$ 0.07} & 90.05 \scriptsize{$\pm$ 0.08} & 87.70 \scriptsize{$\pm$ 0.04} & 94.90 \scriptsize{$\pm$ 0.09} \\ \hline
                DGI                 & 75.35 \scriptsize{$\pm$ 0.14} & 83.95 \scriptsize{$\pm$ 0.47} & 91.61 \scriptsize{$\pm$ 0.22} & 92.15 \scriptsize{$\pm$ 0.63} & 94.51 \scriptsize{$\pm$ 0.52} \\
                GMI                 & 74.85 \scriptsize{$\pm$ 0.08} & 82.21 \scriptsize{$\pm$ 0.31} & 90.68 \scriptsize{$\pm$ 0.17} & OOM & OOM \\
                MVGRL               & 77.52 \scriptsize{$\pm$ 0.08} & 87.52 \scriptsize{$\pm$ 0.11} & 91.74 \scriptsize{$\pm$ 0.07} & 92.11 \scriptsize{$\pm$ 0.12} & 95.33 \scriptsize{$\pm$ 0.03} \\
                GRACE         & \textbf{77.97 \scriptsize{$\pm$ 0.63}} & 86.50 \scriptsize{$\pm$ 0.33} & 92.46 \scriptsize{$\pm$ 0.18} & 92.17 \scriptsize{$\pm$ 0.04} & OOM         \\
                GCA           & 77.94 \scriptsize{$\pm$ 0.67} & 87.32 \scriptsize{$\pm$ 0.50} & 92.39 \scriptsize{$\pm$ 0.33} & 92.84 \scriptsize{$\pm$ 0.15} & OOM         \\ 
                BGRL          & 76.86 \scriptsize{$\pm$ 0.74} & 89.69 \scriptsize{$\pm$ 0.37} & 93.07 \scriptsize{$\pm$ 0.38} & 92.59 \scriptsize{$\pm$ 0.14} & 95.48 \scriptsize{$\pm$ 0.08} \\ \hline
                \proposed    & 77.62 \scriptsize{$\pm$ 0.49} & \textbf{89.88 \scriptsize{$\pm$ 0.33}} & \textbf{93.22 \scriptsize{$\pm$ 0.28}} & \textbf{93.27 \scriptsize{$\pm$ 0.17}} & \textbf{95.69 \scriptsize{$\pm$ 0.10}} \\ 
        \end{tabular}}
    \vspace{-1ex}
            \caption{Performance on node classification (OOM: Out of memory on 24GB RTX3090).}
        \label{tab:main_result}
    }\end{minipage}
    \begin{minipage}{.32\linewidth}{
    	\centering
		\includegraphics[width=0.9\columnwidth]{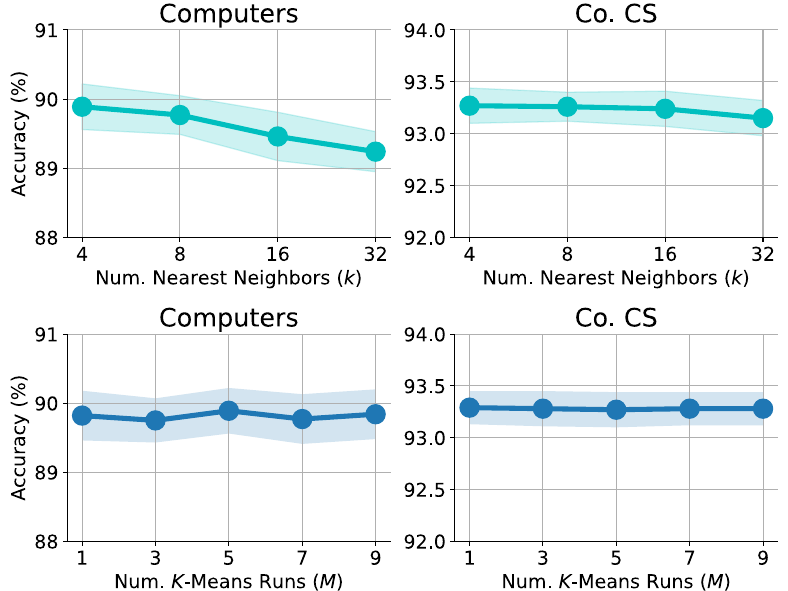}
		\vspace{-1ex}
		\captionof{figure}{Sensitivity analysis.}
	\label{fig:sensitivity} 
	}\end{minipage}
\hfill
\vspace{-3ex}
\end{table*}


\section{Experiments}
\subsection{Experimental Setup}
\subsubsection{Datasets.}
To evaluate~\proposed,
we conduct experiments on five widely used datasets, including WikiCS, Amazon-Computers (\textit{Computers}), Amazon-Photo (\textit{Photo}), Coauthor-CS (\textit{Co.CS}), and Coauthor-Physics (\textit{Co.Physics}). 
\subsubsection{Methods Compared.} 
We primarily compare \proposed~against~GRACE~\cite{GRACE}, BGRL~\cite{BGRL} and GCA~\cite{GCA}, which are the current state-of-the-art self-supervised methods for learning representations of nodes in a graph. 
For all baselines but BGRL, we use the official code published by the authors. As the official code for BGRL is not available, we implement it by ourselves, and try our best to reflect the details provided in the original paper~\cite{BGRL}. 
We also report previously published results of other representative methods, such as DeepWalk \cite{deepwalk}, DGI \cite{DGI}, GMI \cite{GMI}, and MVGRL \cite{MVGRL}, as done in \cite{BGRL,GCA}. 

\subsubsection{Evaluation protocol.}
We evaluate~\proposed~on three node-level tasks, i.e., node classification, node clustering and node similarity search.
We first train all models in an unsupervised manner. 
For node classification, we use the learned embeddings to train and test a simple logistic regression classifier~\cite{DGI}. We report the test performance when the performance on validation data gives the best result. For node clustering and similarity search, we perform evaluations on the learned embeddings at every epoch and report the best performance. 


\subsubsection{Implementation details.}
We use a GCN~\cite{GCN} model as the encoders, i.e., $f_\theta(\cdot)$ and $f_\xi(\cdot)$. 
More formally, the encoder architecture is defined as:
\begin{eqnarray}
   \mathbf{H}^{(l)}= \mathrm{GCN}^{(l)}\mathbf{(X, A)} = \sigma(\mathbf{\hat{D}}^{-1/2}\mathbf{\hat{A}}\mathbf{\hat{D}}^{-1/2}\mathbf{XW}^{(l)}) ,
\end{eqnarray}
where $\mathbf{H}^{(l)}$ is the node embedding matrix of the $l$-th layer for $l\in[1,...,L]$, $\hat{\mathbf{A}} = \mathbf{A} + \mathbf{I}$ is the adjacency matrix with self-loops, $\hat{\mathbf{D}} = \sum_{i}{\hat{\mathbf{{A}}}_{i}}$ is the degree matrix, $\sigma(\cdot)$ is a nonlinear activation function such as ReLU, and $\mathbf{W}^{(l)}$ is the trainable weight matrix for the $l$-th layer. 
We perform grid-search on several hyperparameters, such as learning rate $\eta$, decay rate $\tau$, node embedding dimension size $D$, number of layers of GCN encoder $L$, for fair comparisons.




\begin{table}[t]
	\centering
	\small
	\renewcommand{\arraystretch}{0.8}
	\begin{tabular}{p{1.3cm}|p{0.8cm}|ccc|c}
		\multicolumn{2}{c|}{} & GRACE & GCA & BGRL & \proposed \\ \hline \hline
		\multirow{2}{*}{WikiCS} & NMI & \textbf{0.4282} & 0.3373 & 0.3969 & 0.4132 \\
		& Hom. & \textbf{0.4423} & 0.3525 & 0.4156 & 0.4307 \\ \hline
		\multirow{2}{*}{Computers} & NMI & 0.4793 & 0.5278 & 0.5364 & \textbf{0.5520} \\
		& Hom. & 0.5222 & 0.5816 & 0.5869 & \textbf{0.6040} \\ \hline
		\multirow{2}{*}{Photo} & NMI & 0.6513 & 0.6443 & \textbf{0.6841} & 0.6563 \\
		& Hom. & 0.6657 & 0.6575 & \textbf{0.7004} & 0.6743 \\ \hline
		\multirow{2}{*}{Co.CS} & NMI & 0.7562 & 0.7620 & 0.7732 & \textbf{0.7859} \\
		& Hom. & 0.7909 & 0.7965 & 0.8041 & \textbf{0.8161} \\ \hline
		\multirow{2}{*}{Co.Physics} & NMI & OOM & OOM & 0.5568 & \textbf{0.7289} \\
		& Hom. & OOM & OOM & 0.6018 & \textbf{0.7354} \\
	\end{tabular}
	\vspace{-1.5ex}
	\caption{Performance on node clustering in terms of NMI and homogeneity.}
	\vspace{-3ex}
	\label{tab:clustering}
\end{table}

\begin{table}[t]
	\centering
	\small
	\renewcommand{\arraystretch}{0.9}
	\begin{tabular}{p{1.3cm}|p{1.0cm}|ccc|c}
		\multicolumn{2}{c|}{} & GRACE & GCA & BGRL & \proposed \\ \hline \hline
		\multirow{2}{*}{WikiCS} & Sim@5 & 0.7754 & 0.7786 & 0.7739 & \textbf{0.7811} \\
		& Sim@10 & 0.7645 & \textbf{0.7673} & 0.7617 & 0.7660 \\ \hline
		\multirow{2}{*}{Computers} & Sim@5 & 0.8738 & 0.8826 & 0.8947 & \textbf{0.8966} \\
		& Sim@10 & 0.8643 & 0.8742 & 0.8855 & \textbf{0.8890} \\ \hline
		\multirow{2}{*}{Photo} & Sim@5 & 0.9155 & 0.9112 & \textbf{0.9245} & 0.9236 \\
		& Sim@10 & 0.9106 & 0.9052 & \textbf{0.9195} & 0.9173 \\ \hline
		\multirow{2}{*}{Co.CS} & Sim@5 & 0.9104 & 0.9126 & 0.9112 & \textbf{0.9180} \\
		& Sim@10 & 0.9059 & 0.9100 & 0.9086 & \textbf{0.9142} \\ \hline
		\multirow{2}{*}{Co.Physics} & Sim@5 & OOM & OOM & 0.9504 & \textbf{0.9525} \\
		& Sim@10 & OOM & OOM & 0.9464 & \textbf{0.9486} \\
	\end{tabular}
	\vspace{-1.5ex}
	\caption{Performance on similarity search. (Sim@$n$: Average ratio among $n$ nearest neighbors sharing the same label as the query node.)}
	\vspace{-4ex}
	\label{tab:similarity_search}
\end{table}
\subsection{Performance Analysis}
\subsubsection{Overall evaluation.}
Table~\ref{tab:main_result} shows the node classification performance of various methods. We have the following observations: 
\textbf{1)} Our augmentation-free~\proposed~generally performs well on all datasets compared with augmentation-based methods, i.e., GRACE, GCA and BGRL, whose reported results are obtained by carefully tuning the augmentation hyperparameters. Recall that in Table~\ref{tab:augmentation_sensitivity} we demonstrated that their performance is highly sensitive to the choice of augmentation hyperparameters. This verifies the benefit of our augmentation-free approach. 
\textbf{2)} We also evaluate~\proposed~on node clustering (Table~\ref{tab:clustering}) and similarity search (Table~\ref{tab:similarity_search}). 
Note that the best hyperparameters for node classification task were adopted. 
Table \ref{tab:clustering} shows that~\proposed~generally outperforms other methods in node clustering task. We argue that this is mainly because~\proposed~also considers global semantic information unlike the compared methods. 
\textbf{3)} It is worth noting that methods built upon instance discrimination principle~\cite{IND}, i.e., GRACE and GCA, are not only memory consuming (OOM on large datasets), but also generally perform worse than their counterparts on various tasks (especially on clustering). This indicates that instance discrimination, which treats all other nodes except itself as negatives without considering the graph structural information, is not appropriate for graph-structured data, especially for clustering in which the global structural information is crucial. 
\textbf{4)} ~\proposed~generally performs well on node similarity search (Table~\ref{tab:similarity_search}). This is expected because~\proposed~aims to make nearest neighbors of each node share the same label with the query node by discovering the local and the global positives. 


\begin{figure}[h]
	\centering
	\includegraphics[width=0.93\columnwidth]{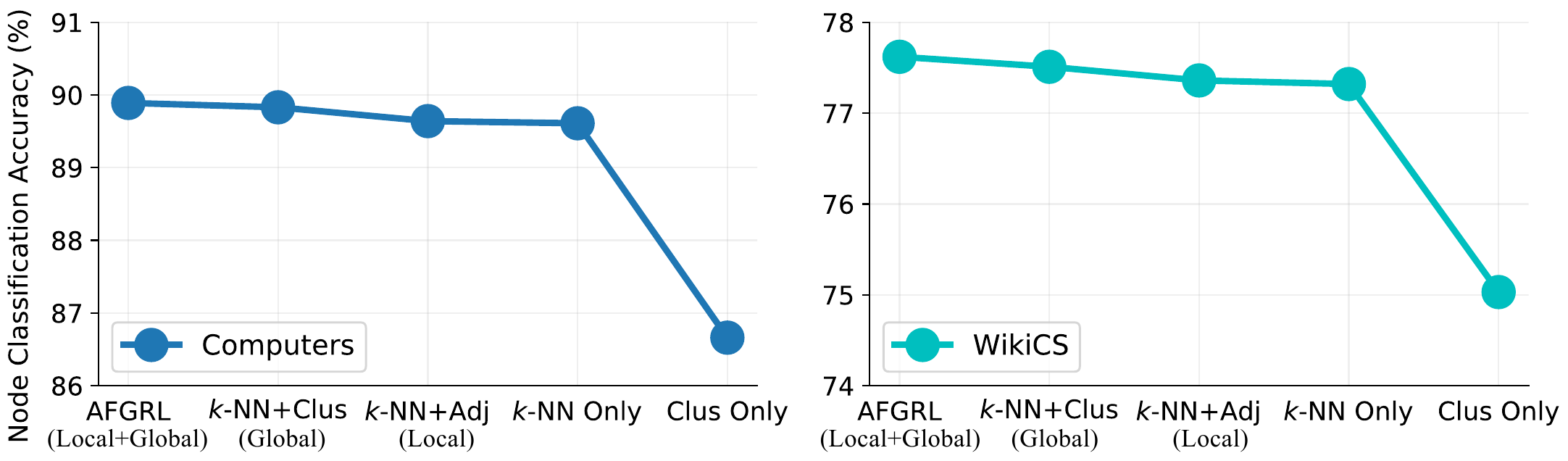} 
	\vspace{-2ex}
	\caption{Ablation study on~\proposed.}
	\label{fig:ablation}
	\vspace{-2ex}
\end{figure}

\subsubsection{Ablation Studies.}
To verify the benefit of each component of~\proposed, we conduct ablation studies on two datasets that exhibit distinct characteristics, i.e., Amazon Computers (E-commerce network) and WikiCS (Reference network). In Figure~\ref{fig:ablation}, we observe that considering both local structural and global semantic information shows the best performance. Moreover, we observe that the global semantic information is more beneficial than the local structural information. This can be explained by the performance of ``\textit{$k$-NN only}'' variant, which performs on par with ``\textit{$k$-NN + Adj}'' variant. That is, we conjecture that performing $k$-NN on the node representations learned by our framework can capture sufficient local structural information contained in the adjacency matrix. Based on the ablation studies, we argue that~\proposed~still gives competitive performance even when the adjacency matrix is sparse, which shows the practicality of our proposed framework. Finally, the low performance of ``\textit{Clus-only}'' variant implies the importance of considering the local structural information in graph-structured data.




\begin{figure}[h]
	\centering
	\includegraphics[width=0.93\columnwidth]{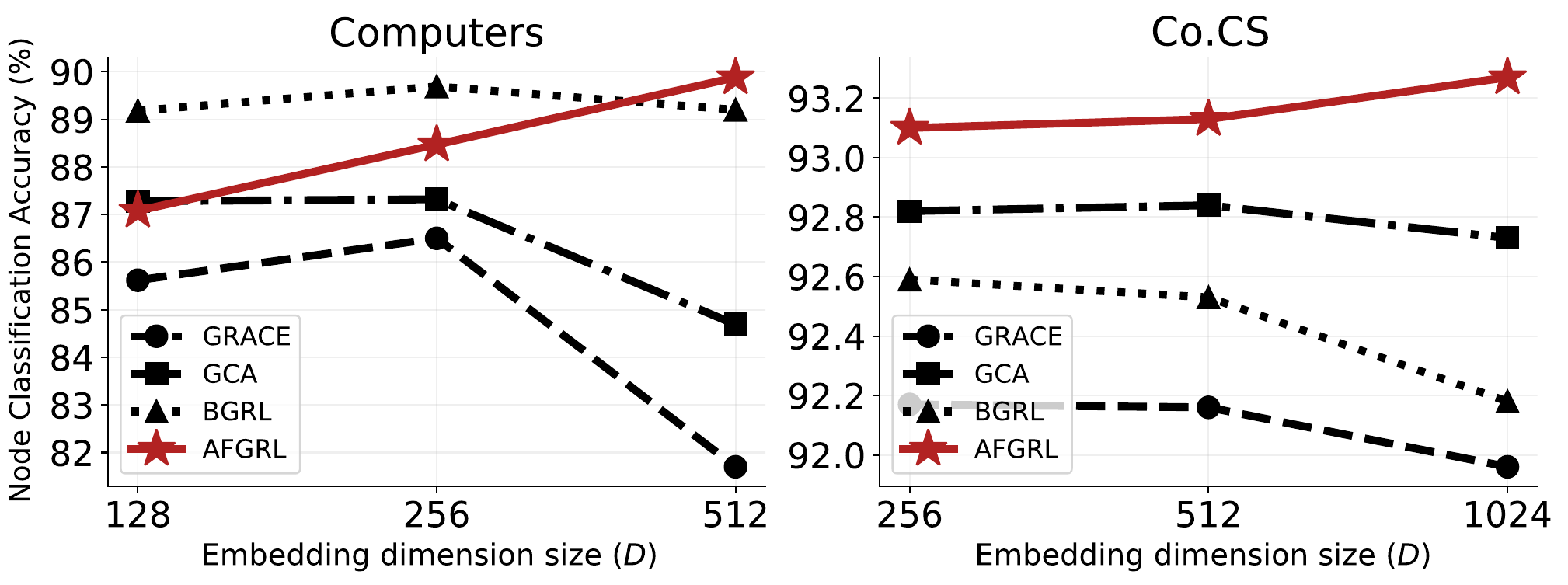}
	\vspace{-1ex}
	\caption{Effect of embedding dimension size ($D$).}
	\label{fig:sensitivitiy_dim}
	\vspace{-2ex}
\end{figure}

\subsubsection{Hyperparameter Analysis.}
Figure~\ref{fig:sensitivity} shows the sensitivity analysis on the hyperparameters $k$ and $M$ of~\proposed. We observe that $k=4$ and $M>1$ generally give the best performance, while the performance is rather stable over various $M$s. 
This verifies that our augmentation-free approach can be easily trained compared with other augmentation-based methods, i.e., stable over hyperparameters, while outperforming them in most cases. 
Moreover, in Figure \ref{fig:sensitivitiy_dim}, we conduct experiments across various sizes of node embedding dimension $D$. We observe that~\proposed~benefits from high-dimensional embeddings, while other methods rapidly saturate when the dimension of embeddings increase. Note that~\citet{zbontar2021barlow} recently showed similar results indicating that methods based on instance discrimination~\cite{IND} is prone to the curse of dimensionality.



\begin{figure}[h]
	\centering
	\includegraphics[width=0.84\columnwidth]{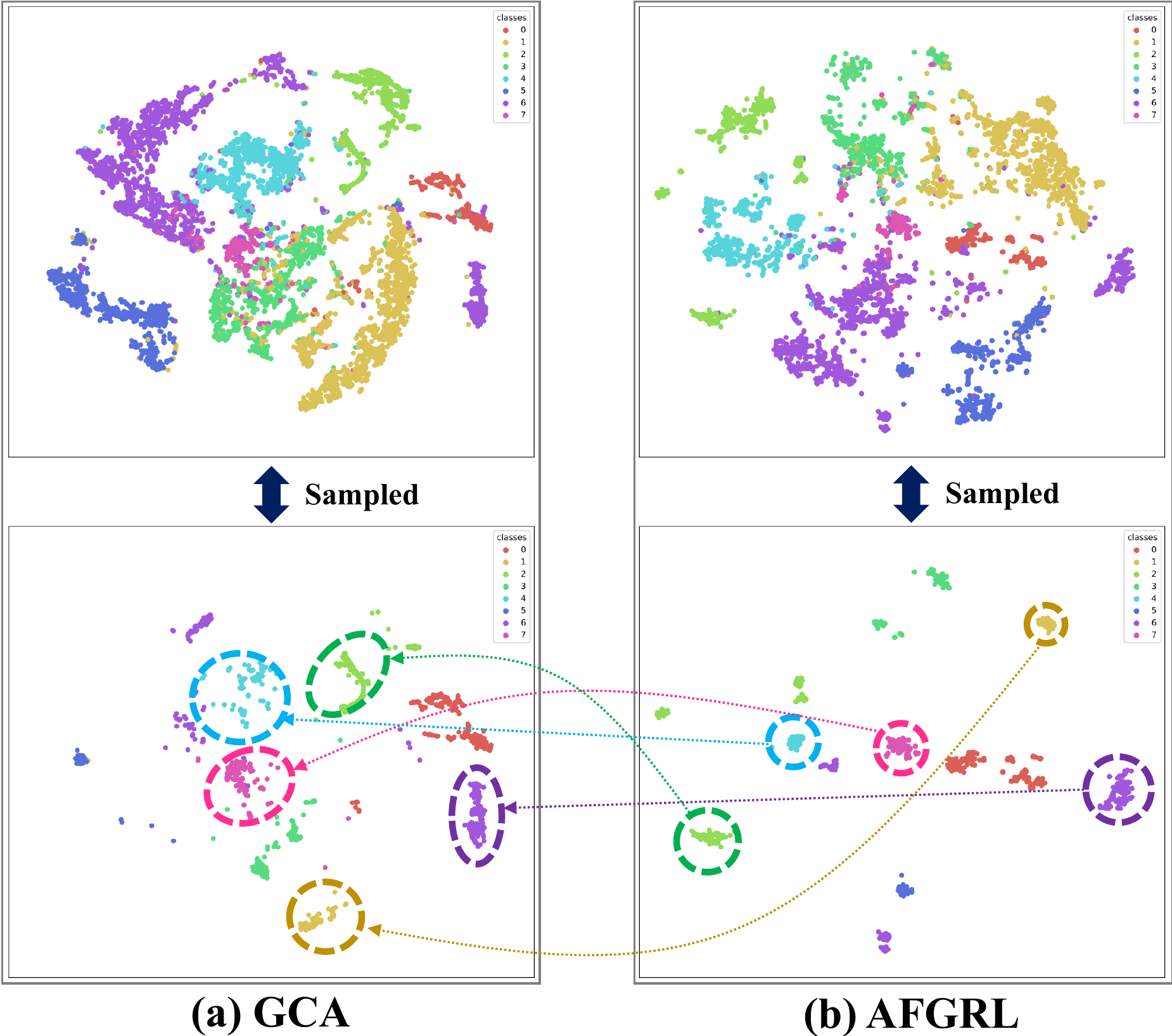}
	\vspace{-1ex}
	\caption{t-SNE embeddings of nodes in \textit{Photo} dataset.}
	\label{fig:tsne}
	\vspace{-2.5ex}
\end{figure}

\subsubsection{Visualization of embeddings.} To provide a more intuitive understanding of the learned node embeddings, we visualize node embeddings of GCA (Figure~\ref{fig:tsne}(a)) and ~\proposed~(Figure~\ref{fig:tsne}(b)) by using t-SNE~\cite{van2008visualizing}. Each point represents a node, and the color represents the node label. We observe that node embeddings generated by both methods are grouped together according to their corresponding node labels. 
However, the major difference is that~\proposed~captures more fine-grained class information compared with GCA. That is, for~\proposed, there tend to be small clusters within each label group. 
To emphasize this, we sample the same set of nodes from each label, and compare their embeddings (Figure~\ref{fig:tsne} bottom). 
We clearly observe that nodes are more tightly grouped in~\proposed~compared with GCA, which implies that~\proposed~captures more fine-grained class information.


\section{Conclusion}
In this paper, we propose a self-supervised learning framework for graphs, which requires neither augmentation techniques nor negative samples for learning representations of graphs.
Instead of creating two arbitrarily augmented views of a graph and expecting them
to preserve the semantics of the original graph,~\proposed~discovers nodes that can serve as positive samples by considering the local structural information and the global semantics of graphs.
The major benefit of~\proposed~over other self-supervised methods on graphs is its stability over hyperparameters while maintaining competitive performance even without using negative samples for the model training, which makes~\proposed~practical.
Through experiments on multiple graphs on various downstream tasks, we empirically
show that~\proposed~is superior to the state-of-the-art methods that are sensitive to augmentation hyperparameters. 

\section{Acknowledgement}
This work was supported by the NRF grant funded by the MSIT (No.2021R1C1C1009081), and the IITP grant funded by the MSIT (No.2019-0-00075, Artificial Intelligence Graduate School Program(KAIST)).

\footnotesize
\bibliography{aaai22}

\clearpage
\section{Appendix}
\subsection{Datasets}
We evaluated the performance of~\proposed~on node-level tasks, i.e., node classification, node clustering, and node similarity search. We conduct experiments on five widely used datasets, including Wiki-CS, Amazon-Computers, Amazon-Photo, Coauthor-CS, and Coauthor-Physics. The detailed statistics are summarized in Table~\ref{tab:data_stats}.
	
\begin{itemize}
\item
\textbf{WikiCS}~\cite{WikiCS} is reference network constructed from Wikipedia references. Nodes denote articles about computer science, and edges denote hyperlinks between the articles. Articles are labeled with 10 related subfields, and their features are average of Glove \cite{glove} embeddings of all words in the article.
\item
\textbf{Amazon-Computers and Amazon-Photo}~\cite{Amazon} are two networks of co-purchase relationships constructed from Amazon. Nodes denote products, and edges exist between nodes if the products are frequently co-purchased. In each dataset, products are labeled with 10 and 8 classes, respectively, based on product category, and the node feature is a bag-of-words representation of words in the product review.
\item
\textbf{Coauthor-CS and Coauthor-Physics}~\cite{MS} are two academic networks containing co-authorship relationship based on Microsoft Academic Graph. Nodes in these graphs denote authors, and edges denote co-authored relationship. In each dataset, authors are classified into 15 and 5 classes, respectively, based on the author's research field, and the node feature is a bag-of-words representation of the paper keywords.
\end{itemize}
For WikiCS dataset, which provides 20 canonical train/valid/test splits, we directly used the given splits. For Amazon and Coauthor datasets, we randomly split nodes 20 times into train/valid/test (10/10/80) as these datasets do not provide standard splits.
	    
\begin{table}[h]
    \centering
	\small
	\begin{tabular}{c|cccc}
	\hline
	& \# Nodes & \# Edges & \# Feat. & \# Cls. \\ \hline \hline
	WikiCS           & 11,701 & 216,123 & 300 & 10         \\
	Amazon-Computers & 13,752 & 245,861 & 767 & 10         \\
	Amazon-Photo     & 7,650 & 119,081 & 745 & 8         \\
	Coauthor-CS      & 18,333 & 81,894 & 6,805 & 15       \\
	Coauthor-Physics & 34,493 & 247,962 & 8,415 & 5         \\ \hline
	\end{tabular}
	\caption{{Statistics for datasets used in this paper.}}
	\label{tab:data_stats}
\end{table}

\subsection{Compared methods}
In this section, we explain methods that are compared with~\proposed~in the experiments, and we summarize their properties in Table~\ref{tab:model_comparison}.
\begin{itemize}
    \item \textbf{DGI}~\cite{DGI}: A pioneering work for self-supervised graph representation learning, which is motivated by Deep InfoMax \cite{DeepInfomax}. DGI aims to learn node representations by maximizing the mutual information between the node and global summary vector of the graph. 
    \item \textbf{GMI}~\cite{GMI}: An advanced version of DGI that learns node representations by leveraging more fine-grained information, i.e. subgraph. Specifically, GMI proposes to directly measure the mutual information between input and node/edge representations within one-hop neighbor, without explicit data augmentation. 
    \item \textbf{MVGRL}~\cite{MVGRL}:  It constructs views of a graph with diffusion kernel and subgraph sampling. Then, it learns to contrast node representations with global summary vector across the two views. 
    \item \textbf{GRACE}~\cite{GRACE}. Inspired by SimCLR \cite{SimCLR}, it first creates two augmented views of a graph by randomly perturbing nodes/edges and their features. Then, following the principle of instance discrimination, it learns node representations by contrasting it with all other nodes in the two augmented graphs, while matching with the same node from the two augmented graphs. 
    \item \textbf{GCA}~\cite{GCA}: An advanced version of GRACE, which proposes multiple augmentation schemes regarding the importance of nodes, edges and their features. The main idea is to selectively augment a graph by keeping important parts of the graph intact, while augmenting unimportant parts.
    \item \textbf{BGRL}~\cite{BGRL}: Inspired by BYOL~\cite{BYOL}, it learns node representations without using negative samples. As conventional contrastive methods, BGRL leverages siamese structured network with augmentation scheme. However, even without relying on negative samples, BGRL prevents a trivial solution through asymmetric model architecture and stop gradient operations.
\end{itemize}
    	
    \begin{table}[h]
        \centering
        \small
        \begin{tabular}{c|cc}
        \hline
        	& \textit{No Augmentation} & \textit{No Negative Sampling}\\\hline \hline
        	DGI     & \xmark & \xmark         \\
        	GMI     & \cmark & \xmark         \\
        	MVGRL   & \xmark & \xmark         \\            
        	GRACE   & \xmark & \xmark         \\
        	GCA     & \xmark & \xmark      \\
        	BGRL    & \xmark & \cmark            \\\hline 
        	\proposed    & \cmark & \cmark          \\ 
        	\hline
        \end{tabular}
        \caption{Properties of methods that are compared.}
        \label{tab:model_comparison}
    \end{table}

\subsection{Implementation details}

\newcolumntype{g}{>{\columncolor{Gray}}c}
	\begin{table*}[t]
    \centering
    \small
    \begin{tabular}{c|ccccccc||ggg}
    \hline
    & \begin{tabular}[x]{@{}c@{}}Embedding\\size ($D$)\end{tabular}    & \begin{tabular}[x]{@{}c@{}}$q_{\theta}$ hidden\\size\end{tabular} & \begin{tabular}[x]{@{}c@{}}Learning\\rate ($\eta$)\end{tabular} & \begin{tabular}[x]{@{}c@{}}Training\\epochs \end{tabular} & Activation &$\tau$ & $L$ & $k$ & $K$ & $M$   \\ \hline\hline
    WikiCS & 1024 & 2048 & 0.001 & 1500 & PReLU & 0.9 & 1 & 8 & 100 & 5   \\
    Amazon Computers & 512 & 1024 & 0.001 & 4000 & PReLU & 0.9 & 1  & 4 & 100 & 5\\
    Amazon Photo & 512 & 1024 & 0.001 & 3000 & PReLU & 0.9 & 1 & 4 & 100 & 5 \\
    Coauthor CS & 1024 & 2048 & 0.001 & 1000 & PReLU & 0.9 & 1 & 4 & 100 & 5 \\
    Coauthor Physics & 256 & 512 & 0.01 & 1000 & PReLU & 0.9 & 1 & 8 & 100 & 5 \\ \hline   
    \end{tabular}
    \caption{Hyperparameter specifications for~\proposed. The three right-most columns denote the hyperparameters that are newly introduced in~\proposed. In contrast to GRACE, GCA and BGRL,~\proposed~does not have hyperparameters associated with graph augmentation.}
    \label{tab:hyperparameters}
\end{table*}
        
As described in Section 6.1 of the submitted manuscript, we use GCN \cite{GCN} encoders. The base encoder of~\proposed~is a GCN model followed by batch normalization and non-linearity. Following BGRL~\cite{BGRL}, the predictor $q_\theta$ of~\proposed~is defined as a multi-layer perceptron (MLP) with batch normalization. 
Note that since a single-layer GCN (i.e., GCN with $L=1$) works the best for~\proposed, the hidden size of $f_\theta$ is not defined. i.e., there is no hidden layer. For GCA and BGRL, we adopt the best hyperparameter specifications that are reported in their original paper, that is, Table 5 of~\cite{GCA} for GCA, and Table 5 of~\cite{BGRL} for BGRL. For GRACE, since the original paper~\cite{GRACE} did not evaluate on the datasets used in our experiments, we follow the best hyperparameter specifications that are reported in the GCA paper~\cite{GCA}, since GRACE is equivalent to the GCA-T-A ablation of GCA, which is also proposed by the same authors.
Refer to Table~\ref{tab:hyperparameters} for more detailed hyperparameter specifications of~\proposed. 

It is important to note that~\proposed~does not have hyperparameters associated with graph augmentation, whose best performing combinations is non-trivial to find\footnote{The values of the best performing hyperparameters (i.e., $p_{f,1}$, $p_{f,2}$, $p_{e,1}$, and $p_{e,2}$) vary greatly in range from 0.1 to 0.5. Refer to Table 5 of~\cite{GCA,BGRL}}. Instead,~\proposed~newly introduced several hyperparameters, i.e., $k$, $K$, and $M$. However, we observe that the model performance is stable over these hyperparameters as shown in Figure 5 of the submitted manuscript and Figure~\ref{exp:ablation_clustering} of this supplementaty material.
Hence, we fixed $K$ to 100, $M$ to 5, and selected $k\in\{4,8\}$. This demonstrates the practicality of~\proposed.

\subsection{Reproducibility}
\begin{table}[h]
	\centering
	\small
	\begin{tabular}{c|c}
	\hline
		Methods & Source code \\ \hline
		GRACE & \url{https://github.com/CRIPAC-DIG/GRACE} \\
		GCA & \url{https://github.com/CRIPAC-DIG/GCA}\\
		BGRL & \url{https://github.com/Namkyeong/BGRL_Pytorch}\\\hline
		\proposed & \url{https://github.com/Namkyeong/AFGRL}\\
		\hline
	\end{tabular}%
	\caption{Source code links of the baseline methods.}
	\label{tab:codes}%
\end{table}%
Table~\ref{tab:codes} shows the github links to the source codes we used for evaluation. All the compared methods but BGRL were publicly available, and thus we implemented BGRL using PyTorch on our framework. 

\end{document}